\documentclass[10pt,twocolumn,letterpaper]{article}

\usepackage{cvpr}              %

\usepackage[dvipsnames]{xcolor}
\usepackage{comment}
\usepackage{float}

\newcommand{\TODO}[1]{}
\newcommand{\camready}[1]{{\color{black} #1}}
\newcommand{\qheading}[1]{\noindent\textbf{#1}}
\newcommand\blfootnote[1]{%
  \begingroup
  \renewcommand\thefootnote{}\footnote{#1}%
  \addtocounter{footnote}{-1}%
  \endgroup
}
\usepackage[accsupp]{axessibility} %
\usepackage[subtle]{savetrees}

\newcommand{\todo}[1]{}
\newcommand{\todorebuttal}[1]{}

\newcommand{\change}[1]{\textcolor{black}{#1}}

\newcommand{\methodtitleshort}{\textit{SCULPT}\xspace}
\newcommand{\dataset}{\textit{SCULPT dataset}\xspace}

\newcommand{\randomnoise}{\mathbf{z}}
\newcommand{\randomnoisegeo}{\mathbf{z}_{g}}
\newcommand{\randomnoisetex}{\mathbf{z}_{t}}

\newcommand{\clothconditiongeo}{\mathbf{c}_{g}}
\newcommand{\clothconditiontex}{\mathbf{c}_{t}}

\newcommand{\texturemap}{\boldsymbol{\mathcal{I}}_{tex}}
\newcommand{\generatorgeo}{\mathcal{G}_{geo}}
\newcommand{\generatortex}{\mathcal{G}_{tex}}
\newcommand{\generatorgeoverts}{V_{geo}}

\newcommand{\dispmap}{UV_{geo}}

\newcommand{\shapecoeff}{\boldsymbol{\beta}}

\newcommand{\posecoeff}{\boldsymbol{\theta}}

\newcommand{\numjoints}{k}
\newcommand{\joints}{\textbf{j}}

\newcommand{\mesh}{\boldsymbol{\mathcal{M}}}

\newcommand{\stdnormal}[1]{\mathcal{N}(\mathbf{0}, I^{#1\times#1})}

\definecolor{cvprblue}{rgb}{0.21,0.49,0.74}
\usepackage[pagebackref,breaklinks,colorlinks,citecolor=cvprblue]{hyperref}

\def\paperID{14190} %
\def\confName{CVPR}
\def\confYear{2024}

\title{SCULPT: Shape-Conditioned Unpaired Learning of Pose-dependent Clothed \\ and Textured Human Meshes}

\author{Soubhik Sanyal$^{1}$ \qquad Partha Ghosh$^{1}$ \qquad Jinlong Yang$^{1*}$ \\ \qquad Michael J. Black$^{1}$ \qquad Justus Thies$^{1,2}$ \qquad Timo Bolkart$^{1*}$ \\
\textrm{$^1$Max Planck Institute for Intelligent Systems} \quad \textrm{$^2$Technical University of Darmstadt} \\
{\tt\small \{soubhik.sanyal, partha.ghosh, jinlong.yang, black, justus.thies, timo.bolkart\}@tue.mpg.de}
}

\begin{document}

\twocolumn[{%
	\renewcommand\twocolumn[1][]{#1}%
	\maketitle
	\begin{center}
		\centerline{
                \includegraphics[width=\linewidth, trim={16.5cm 14.5cm 16cm 14.5cm},clip]{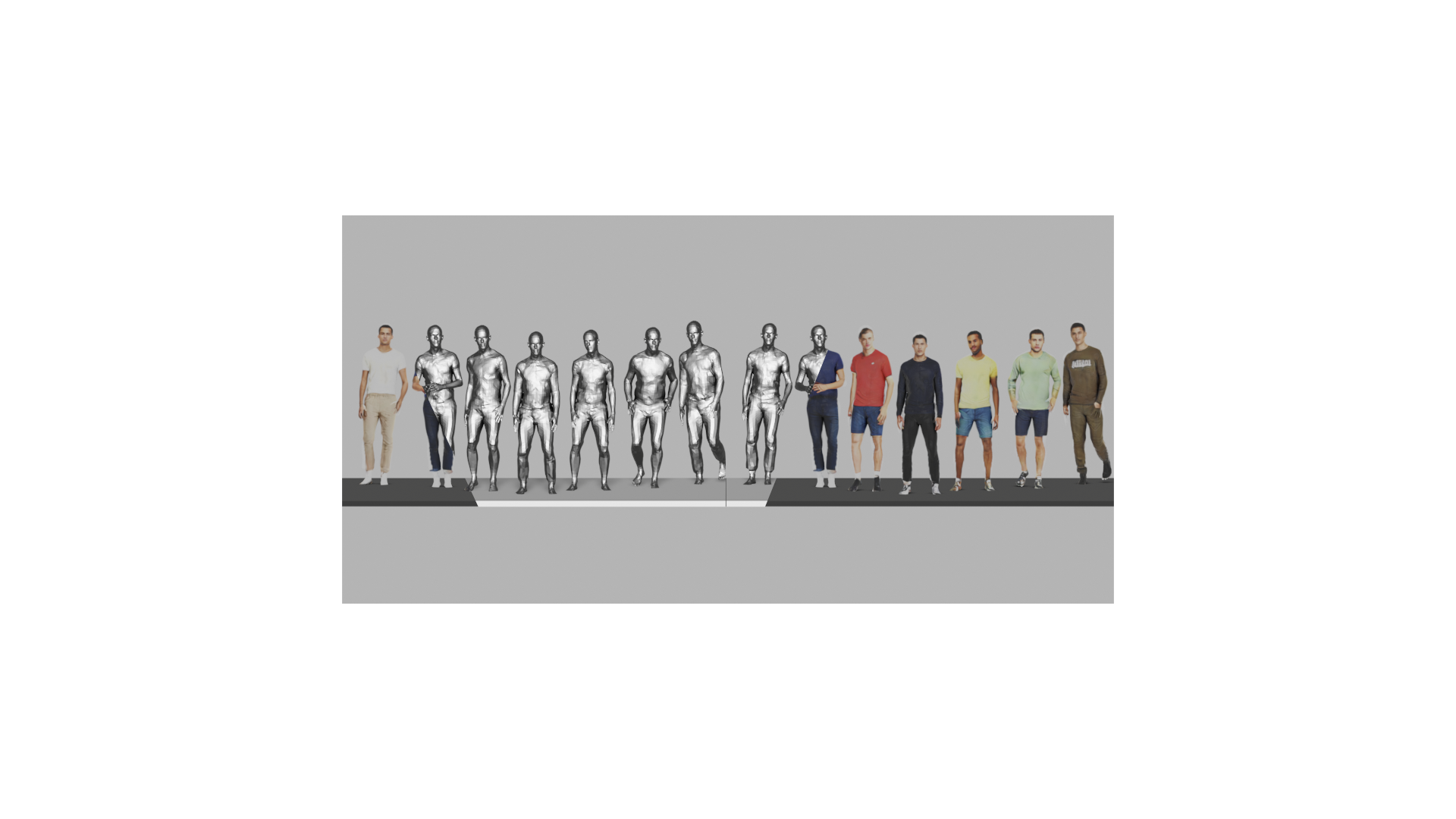}
		}
        \captionof{figure}{
        \textbf{SCULPT} is a generative model of geometry and appearance of clothed human meshes. 
        The generated textured clothing mesh can be readily inserted into 3D scenes. 
        In the above figure, we generated the clothed humans and placed them on a 3D floor. 
        The scene has a single camera with global directional light settings.}
        \label{fig:teaser}
	\end{center}
}]

\maketitle
\blfootnote{*Now at Google.}
\begin{abstract}
We present \methodtitleshort, a novel 3D generative model for clothed and textured 3D meshes of humans.
Specifically, we devise a deep neural network that learns to represent the geometry and appearance distribution of clothed human bodies.
Training such a model is challenging, as datasets of textured 3D meshes for humans are limited in size and accessibility.
Our key observation is that there exist medium-sized 3D scan datasets like CAPE, as well as large-scale 2D image datasets of clothed humans and multiple appearances can be mapped to a single geometry.
To effectively learn from the two data modalities, we propose an unpaired learning procedure for pose-dependent clothed and textured human meshes.
Specifically, we learn a pose-dependent geometry space from 3D scan data. We represent this as per vertex displacements w.r.t. the SMPL model.
Next, we train a geometry conditioned texture generator in an unsupervised way using the 2D image data.
We use intermediate activations of the learned geometry model to condition our texture generator. 
To alleviate entanglement between pose and clothing type, and pose and clothing appearance, we condition both the texture and geometry generators with attribute labels such as clothing types for the geometry, and clothing colors for the texture generator. 
We automatically generated these conditioning labels for the 2D images based on the visual question answering model BLIP and CLIP. %
We validate our method on the \dataset, and compare to state-of-the-art 3D generative models for clothed human bodies.
\camready{Our code and data can be found at \url{https://sculpt.is.tue.mpg.de}.}
\end{abstract}
    
\section{Introduction}
\label{sec:intro}

Generating 3D virtual humans that can be articulated with realistically deforming clothing, is a key challenge of content creation for games and movies.
Virtual assistants in augmented and virtual reality could be enriched by an automatically generated 3D human appearance.
Furthermore, synthetic humans \camready{could} also play an important role in data generation to comply with privacy and data protection laws.

Recently, we have seen immense progress in the synthesis of virtual 3D humans~\cite{hong2023eva3d, zhang2022avatargen, grigorev2021stylepeople, bergman2022generative, noguchi2022unsupervised, Jiang2022_HumanGen}.
However, they are almost exclusively based on implicit representations, such as neural radiance fields~\cite{noguchi2022unsupervised, hong2023eva3d}.
These implicit representations are incompatible with classical rendering frameworks, making it challenging to integrate them into existing applications.
\camready{To address this}, we propose \methodtitleshort a generative model \camready{of} explicit 3D geometry (meshes) and the appearance (texture maps) of clothed humans (Fig.~\ref{fig:teaser}).
There are several challenges that we need to address to build such a model.
To naively train the 3D generative model, a large dataset of 3D scanned humans would be required containing a diverse set of people in a variety of different poses, wearing different outfits.
Not only such data is not publicly available, but it would also be very expensive to collect.
However, we observed that there are different requirements for learning the geometry and texture of a generative model.
\camready{
For geometry learning, there are medium-scale datasets of 3D-scanned humans that are publicly accessible. 
Additionally, 2D images depicting humans in various outfits are abundantly available, which can assist in learning appearance or texture. 
It is also noted that clothing items, such as t-shirts, despite having identical geometries, can vary significantly in colors and patterns. 
This diversity allows for a wide range of colors and textures to be associated with similar geometrical structures.}
Based on this observation, we propose to leverage medium-scale datasets of 3D scans to learn the geometry distribution, while large-scale 2D image datasets are used to learn the appearance model.
Based on these data sources, and leveraging the foundational SMPL body model~\cite{SMPL:2015}, we design our explicit generative model.

\camready{Specifically,} we modify the StyleGAN architecture to output (i) pose-dependent geometry in terms of displacement maps, and (ii) geometry-dependent appearance in terms of a color texture, in the texture space (UV-space) of the SMPL template.
Our geometry model is trained using the CAPE dataset~\cite{ma2020cape}, which contains pose annotations as well as registered SMPL meshes.
We use the trained geometry model to condition our texture generator, which is trained on a large collection of 2D fashion images.
Note that we train the texture model in an unsupervised way, only relying on adversarial losses, thereby avoiding the requirement of 3D data paired with 2D images. 
The process of geometry conditioning plays a crucial role in maintaining coherence between appearance and geometry. In essence, this process ensures that the visual and geometric attributes of the clothed human harmoniously conform to each other.
To mitigate dependence of generated clothing type or its color on the body pose, we condition both the texture and the geometry generators with attribute labels. 
This approach reduces the entanglement between pose and clothing attributes, resulting in more realistic and accurate generation.
We automatically generate these labels by using the visual question-answering model BLIP~\cite{li2022blip}, and CLIP~\cite{radford2021learning}.

During inference, our model has the ability to generate a diverse set of clothed, textured 3D humans that can be controlled by various parameters such as clothing type, and clothing appearance such as color.
This level of control over the generated output is a significant advantage of our approach, as it allows for great flexibility in generating realistically clothed and textured 3D humans. 
The coarse clothing color can be controlled from text-based inputs, which enables users to specify the desired color for a given piece of clothing without requiring detailed knowledge of 3D modeling or design.
Moreover, since our generative model produces 3D meshes it is compatible with existing graphics and game engines, allowing seamless integration into a range of applications. 
In summary, the contribution of this work is a novel hybrid learning strategy for unsupervised and unpaired learning of a generative model of 3D virtual humans. It is enabled by coupling the appearance and geometry network via language-driven attribute labels.

\section{Related work}

\begin{figure*}[t]
	\centerline{
		\includegraphics[width=1.0\linewidth, trim={0cm, 6cm, 0cm, 5cm}, clip]{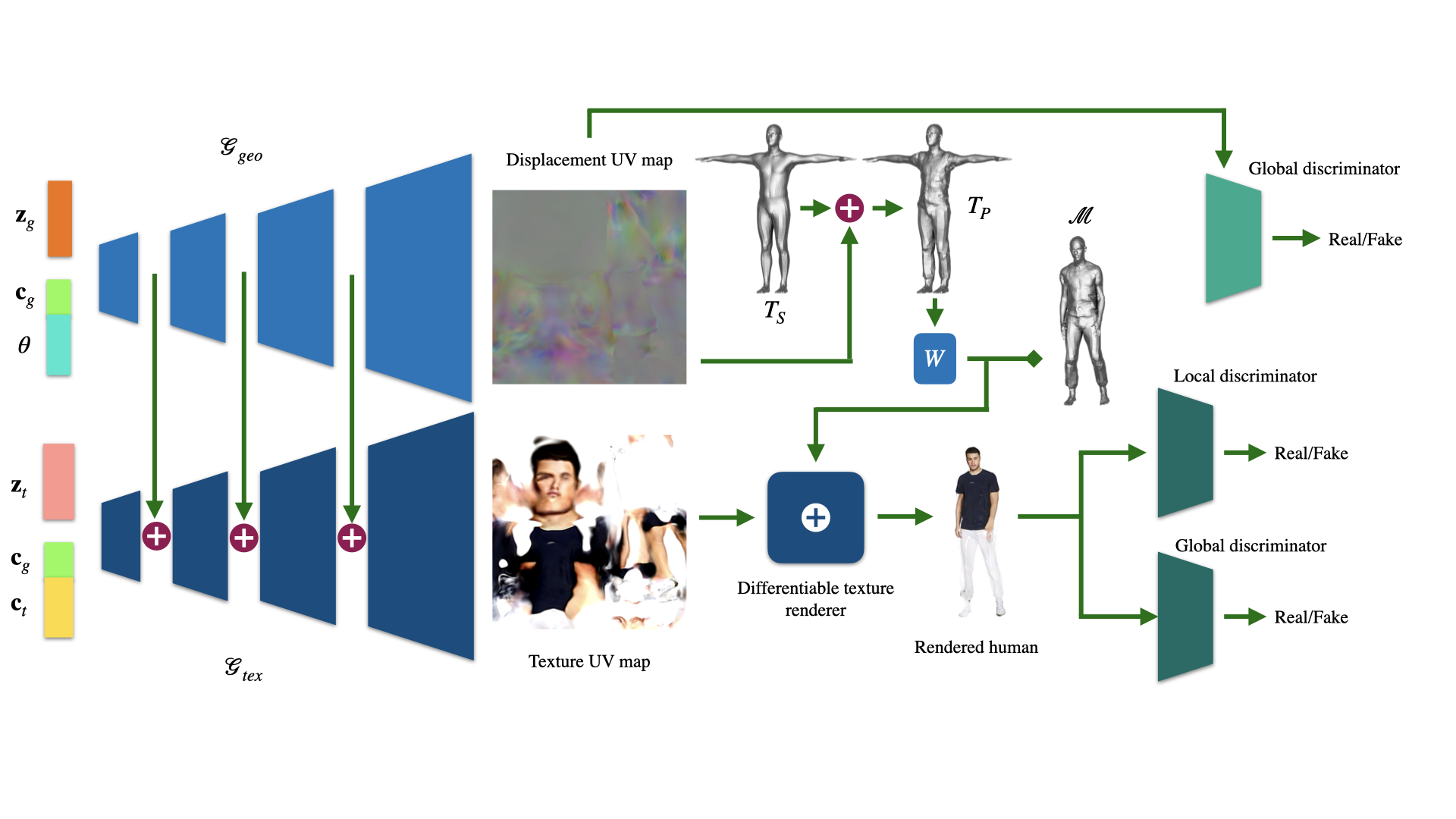}
	}
	\caption{
    \textbf{\camready{Overview:}} 
    SCULPT consists of two StyleGAN-based generators for geometry ($\generatorgeo$) and appearance ($\generatortex$), both acting in the UV space of the SMPL body model. 
    The geometry network $\generatorgeo$ outputs pose-dependent displacement maps that are added to the SMPL template mesh and is trained using 3D scan data.
    Based on this model, the appearance generator $\generatortex$ is trained in an unsupervised way using adversarial losses computed on rendered images of the generated synthetic human.
    It is conditioned on intermediate features of the geometry network.
    Besides the noise code, both generator networks receive additional attributes for appearance ($\clothconditiontex$) and clothing type ($\clothconditiongeo$) as input.
    This enhances the connection between appearance and geometry, %
    and it offers a user-friendly control over the generation. 
    \TODO{add symbols for the networks etc. to the caption text. Done}
	}
    \label{fig:dispmap_2_mesh}
\end{figure*}

\paragraph{Generative 3D modeling:}
3D-aware generative models have been receiving a tremendous amount of focus from the computer vision community in recent years~\cite{Chan2022, orel2022stylesdf, gao2022get3d, chan2021pi, schwarz2020graf, schwarz2022voxgraf}. 
\camready{This has resulted in different representations of 3D objects and scenes like implicit functions and different rendering techniques such as volumetric rendering.} 
In the following paragraph, we provide a brief overview of these concepts as they help contextualize our work better.

Chan~\etal~\cite{chan2021pi} propose a new architecture for generative models, where they build the generator with SIREN networks~\cite{sitzmann2020implicit} and perform  volumetric rendering to generate 2D images.
The SIREN network models the inherent geometry of an image using implicit representations which are rendered to a 2D image via volumetric rendering inspired by NeRF~\cite{mildenhall2020nerf}.
Or-El~\etal~\cite{orel2022stylesdf} combine an SDF-based 3D representation with a style-based 2D generator. 
The SDF-based representation helps in achieving geometry details but produces low-resolution image features, which are then fed to a 2D-based style generator that produces high-resolution images.
Chan \etal \cite{Chan2022} take an interesting direction by decoupling feature generation and neural rendering with the help of \camready{a} tri-plane representation.
This enables them to leverage the power of superior image generation modules like StyleGAN2~\cite{karras2020analyzing} to generate high-quality 2D results.
Even though these works have \camready{advanced} the state-of-the-art in 3D aware image synthesis for relatively simple objects like cars or faces, their generated geometry is not of high quality.
Moreover, they are not directly usable for complex 3D deformable and articulable models like clothed humans.

This leads to a more focused set of generative models~\cite{hong2023eva3d, zhang2022avatargen, grigorev2021stylepeople, bergman2022generative, noguchi2022unsupervised, Jiang2022_HumanGen, ma2020cape, chen2022gdna, corona2021smplicit, palafox2021npms} which are concentrated on generative 3D humans.
Among these 3D generative models, a stream of work~\cite{ma2020cape, chen2022gdna, corona2021smplicit, palafox2021npms} uses 3D data as supervision. 
\camready{Yet they only model geometry and not the texture or appearance of the clothed humans.
This is caused by the limited size of 3D datasets containing the large variation of clothing textures along with their corresponding geometry.}
To get around \camready{this limitation}, another stream of work uses 2D images as supervision.
The concurrent works by Bergman et al.~\cite{bergman2022generative}, Zhang et  al.~\cite{zhang2022avatargen} and Noguchi et al.~\cite{noguchi2022unsupervised} extend Chan et al.~\cite{Chan2022} for human bodies where they learn a neural radiance field for a canonical pose \camready{that} is later reposed.
The  radiance fields are learned by using a similar concept of tri-planes proposed by Chan et al.~\cite{Chan2022}.
The final human image produced by the network of~\cite{bergman2022generative} is blurry however, and the geometry is very smooth and lacks deformations relating to clothing.
Similarly, Noguchi \etal~\cite{noguchi2022unsupervised} also produce blurry results and undesired artifacts.
\change{
Grigorev \etal \cite{grigorev2021stylepeople} propose to generate neural textures using 2D generative models.
The neural texture is superimposed on minimally clothed SMPL-X~\cite{SMPL-X:2019} meshes and rendered.
The rendered images are then passed via a neural renderer to generate realistic-looking 2D person images.
Their method does not alter the geometry and is incompatible with existing infrastructure, \eg, game engines.}
Hong \etal \cite{hong2023eva3d} divide the whole body into parts and learn local radiance fields, which are then integrated to get the final rendered image. 
They also suffer from undesirable artifacts and lack any control over geometry and texture.
In contrast, we learn geometry on top of a fixed topology mesh (SMPL,~\cite{SMPL:2015}) and the corresponding texture map for the underlying mesh.
This is done via learning from unpaired data using a novel training approach utilizing both 3D and 2D data.
\camready{Displacements on top of SMPL are sufficient for many kinds of clothes. %
Moreover, this template} is readily compatible with any existing 3D rendering engine.
Additionally, unlike prior art, we provide explicit controls over the clothing type and color. 

\paragraph{Generative 2D humans:}
If we set aside the goal of having an explicit 3D mesh as output, some 2D methods become relevant. 
Here, we study a few generative models that can generate high-quality 2D images for clothed humans~\cite{Sarkar2021_HumanGAN, Fu2022_StyleGANHuman}. 
Fu \etal \cite{Fu2022_StyleGANHuman} propose a curated dataset of fashion images and train StyleGAN~\cite{Karras2021, karras2020analyzing}.
\camready{In this way,} they produce high-quality 2D images of clothed humans but lack controllability on the generated pose and other aspects, such as global orientation, clothing type\camready{, etc.} %
Sarkar \etal \cite{Sarkar2021_HumanGAN} propose a new architecture for controllable human generation. 
To provide additional pose controllability over the generated humans one can use any of the 2D reposing algorithms~\cite{balakrishnan2018synthesizing, dong2018soft, dong2020fashion, grigorev2019coordinate, knoche2020reposing, esser2018variational, pumarola2018unsupervised, yang2020towards, sanyal2021learning, ma2018disentangled, pumarola2018unsupervised, sanyal2021learning, song2019unsupervised, Sarkar2021, AlBahar2021} proposed in the recent years.
In summary, these methods propose various architectures and/or training procedures that enable them in synthesizing a human in a new pose given the image of the person in a different pose.
\camready{Although these methods generate} high-quality images, they have no notion of the underlying geometry.
Therefore they cannot be used in classical 3D graphics pipelines.%

\section{Method}

\methodtitleshort is a generative model \camready{that} takes a geometry code $\randomnoisegeo\sim\stdnormal{512}$, a texture code $\randomnoisetex\sim\stdnormal{512}$, body pose $\posecoeff\in\mathbb{R}^{69}$, clothing geometry type $\clothconditiongeo\in \{0, 1\}^{6}$, and clothing texture description $\clothconditiontex\in\mathbb{R}^{512}$ as input, and it generates a clothed 3D body mesh $\mesh := \{V, \mathcal{C}\}$ with a texture image $\texturemap \in \mathbb{R}^{256\times256\times3}$~(\cref{fig:dispmap_2_mesh}). 
Here, $V \in \mathbb{R}^{6890\times3}$ represents a set of $6890$ 3D vertices, and $\mathcal{C}$ is a fixed set of triangles represented as a 3-tuple of vertex indices in SMPL~\cite{SMPL:2015} mesh topology. 
Formally, \methodtitleshort is defined as:
\begin{equation}
    \mesh, \texturemap = \mathit{SCULPT}(\randomnoisegeo, \randomnoisetex, \posecoeff,
    \clothconditiongeo, \clothconditiontex).
\end{equation}
\methodtitleshort models clothing geometry as vertex displacements from the minimally clothed SMPL body in the canonical pose. 
To account for pose articulation effects, the clothing generator is conditioned on pose.
The texture generator in \camready{turn} takes the features from the geometry generator as a conditioning signal.
Therefore, the output of \methodtitleshort can be fully articulated with SMPL's pose control, and the mesh is readily usable in existing graphics applications.

\subsection{Clothing representation}
\label{sec:clothing_representation}

\methodtitleshort adapts the SMPL~\cite{SMPL:2015} model formulation to clothed bodies with additional parameters $\randomnoisegeo$ and $\clothconditiongeo$ as:
\begin{equation}
    T_P(\shapecoeff, \posecoeff, \randomnoisegeo, \clothconditiongeo) = T_S(\shapecoeff) + B_P(\posecoeff) +  \generatorgeoverts(\randomnoisegeo,\clothconditiongeo,\posecoeff),
    \label{eq:clothed_shape}
\end{equation}
where $T_S(\shapecoeff) \in \mathbb{R}^{6890\times3}$ denotes the SMPL body in ``canonical pose'' for the shape parameters \camready{$\shapecoeff \in \mathbb{R}^{10}$}.  %
\camready{$B_P(\posecoeff): \mathbb{R}^{3\numjoints} \rightarrow \mathbb{R}^{6890\times3}$} are the SMPL pose corrective blend shapes, and $\generatorgeoverts$ are the pose-dependent clothing vertex displacements. 
The vertex displacements $\generatorgeoverts$ are obtained by sampling the UV displacement map output by the clothing geometry generator $\generatorgeo$ (see \cref{sec:geometry_generator}) at fixed UV coordinates for every vertex. 

As the clothing geometry is defined as offsets from the SMPL body in the canonical pose, it can be reposed as:
\begin{equation}
    \mesh(\shapecoeff, \posecoeff, \randomnoisegeo, \clothconditiongeo) = W(T_P(\shapecoeff, \posecoeff, \randomnoisegeo, \clothconditiongeo), \joints(\shapecoeff), \posecoeff),
    \label{eq:final_mesh}
\end{equation}
\camready{where $W(T_P, \joints, \posecoeff)$ is SMPL's blend skinning function, which rotates the vertices of $T_P$ around $\numjoints = 23$ joints $\joints \in \mathbb{R}^{3\numjoints}$.
The joint locations $\joints$ are defined as a function of the body shape.}

\subsection{Clothing geometry generator}
\label{sec:geometry_generator}

\begin{figure}[t]
    \centering
	\centerline{
		\includegraphics[width=\linewidth, trim={9cm 10cm 10cm 10cm},clip]{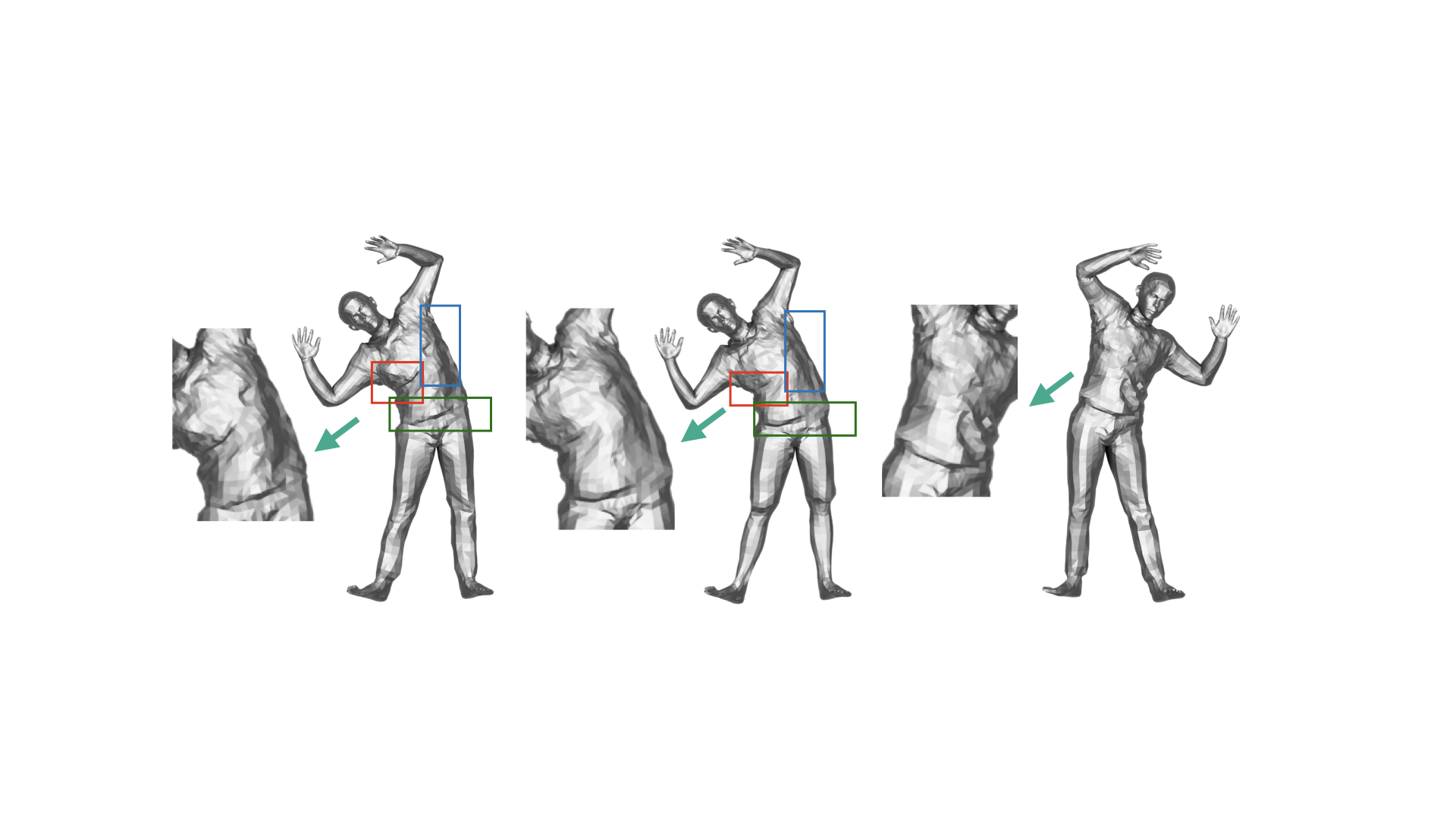}
	}
        \caption{\camready{\textbf{Pose control:} The figure presents three meshes of an individual. The initial two meshes depict the individual in different types of clothing (long-long and short-short) but maintaining the same pose, while the third mesh illustrates the individual in a different pose, wearing the same type of clothing as depicted in the first mesh. The pose-dependent clothing deformations are visible in the zoomed-in images on the side, which are produced by the geometry generator. It is observed that identical poses result in similar deformations, as indicated by the color-coded bounding boxes, while a change in pose leads to distinct clothing deformations in the geometry.}
        }
    \label{fig:pose_dependent_deformations}
\end{figure}

Given \camready{a random} geometry code $\randomnoisegeo\sim \stdnormal{512}$, a one-hot clothing type vector $\clothconditiongeo \in \{0, 1\}^6$, and the SMPL body pose $\posecoeff$, the clothing geometry generator outputs a UV displacement map $\dispmap \in \mathbb{R}^{256\times256\times3}$ (see left column of \cref{fig:dispmap_2_mesh}).
Formally, the generator is defined as:
\begin{equation}
    \dispmap = \generatorgeo(\randomnoisegeo|\clothconditiongeo, \posecoeff).
\end{equation}
Following CAPE~(\cite{ma2020cape}), $\clothconditiongeo$ are categorical classifications of the clothing type like
\camready{``short sleeve T-shirt/short trouser'', ``short sleeve T-shirt/long trouser'', ``long sleeve T-shirt/long trouser'', ``long sleeve T-shirt/short trouser'', ``shirt/long trouser'', ``shirt/short trouser''}, provided with the CAPE dataset, and which are represented as one-hot vectors. 
Intuitively, $\clothconditiongeo$ controls the clothing category, while $\randomnoisegeo$ and $\posecoeff$ model clothing variations and pose-dependent deformations within the particular clothing category, respectively~(\cref{fig:pose_dependent_deformations}). 
To sample clothed 3D body meshes from \methodtitleshort, we sample vertex displacements $\generatorgeoverts$ from the generated displacement map $\dispmap$ and evaluate \cref{eq:final_mesh}.

The generator $\generatorgeo$ follows a StyleGAN3~\cite{Karras2021} architecture, which is trained with an adversarial loss from a global discriminator. 
We compute the displacement maps from the scan registrations provided by the CAPE dataset, which are treated as real samples for the discriminator.
For the fake examples, we combine random noise vectors for $\randomnoisegeo$ and $\clothconditiongeo$ with randomly sampled CAPE data poses. 
During training, the output of the generator is masked by the segmentation mask of the SMPL UV map and passed to the discriminator as fake samples.
Following standard conditional GAN training, the discriminator is also given $\clothconditiongeo$ and $\posecoeff$ as inputs.

\subsection{Texture generator}
\label{sec:texture_generator}

Given a random texture code $\randomnoisetex\sim \stdnormal{512}$, our texture generator $\generatortex$ generates a UV texture image. 
To control the generated texture $\generatortex$, it is conditioned on clothing texture descriptors $\clothconditiontex$.
Additionally we condition the texture generator with the categorical clothing type $\clothconditiongeo$ and intermediate features of the clothing geometry generator $\generatorgeo$.
Formally, the texture generator is:
\begin{equation}
    \texturemap = \generatortex(\randomnoisetex|\clothconditiongeo, \clothconditiontex, \generatorgeo).
\end{equation}

We train the texture generator from a collection of 2D fashion images, obtained from fashion websites. 
For each training image, as detailed in \cref{sec:dataprocessing}, we automatically extract clothing color descriptors $\clothconditiontex$ and clothing types $\clothconditiongeo$ with the help of a visual question answering system -- BLIP~(\cite{li2022blip}) and CLIP~(\cite{radford2021learning}).
Realistic clothing appearance, however, consists of more than coarse clothing colors. Namely, it can contain varying color patterns. We capture such variations in our generator using an additional latent vector $\randomnoisetex$. 
The coarse clothing geometry type $\clothconditiongeo$ and the clothing color descriptor $\clothconditiontex$ are concatenated, and provided as input condition to $\generatortex$.
This is sufficient to generate good visual quality texture but the generated texture is inconsistent with the generated geometry. For instance, a short-sleeved shirt covers a smaller skin region of the arms as compared to a long-sleeved one, therefore the texture generator must be conditioned using the geometry information.
Although the clothing category $\clothconditiongeo$ loosely \camready{correlates with} the generated texture and geometry, it is not enough. For instance,  
without explicit information on the clothing part boundaries (e.g. boundary between shirt and trousers, or boundary between cloth and skin) the texture network is unable to generate a clothing appearance that conforms to the clothing geometry. 
To account for the correlation of generated geometry and texture, we condition $\generatortex$ on intermediate features of $\generatorgeo$.
Specifically, during a forward pass, $\generatorgeo$ takes the batch of $\randomnoisegeo, \clothconditiongeo$, and $\posecoeff$, and it generates features at different synthesis blocks.
Since $\generatortex$ and $\generatorgeo$ share StyleGAN's model architecture, we progressively add at each level,  excluding the mapping network, the feature blocks of $\generatorgeo$ to the feature blocks of $\generatortex$. See the left half of~\cref{fig:dispmap_2_mesh} for a visual representation of this technique.
This effectively passes the signals from the geometry network to the texture generator.

\begin{figure}[t]
    \centering
	\centerline{
		\includegraphics[width=\linewidth, trim={7cm, 2.0cm, 4cm, 2.0cm}, clip]{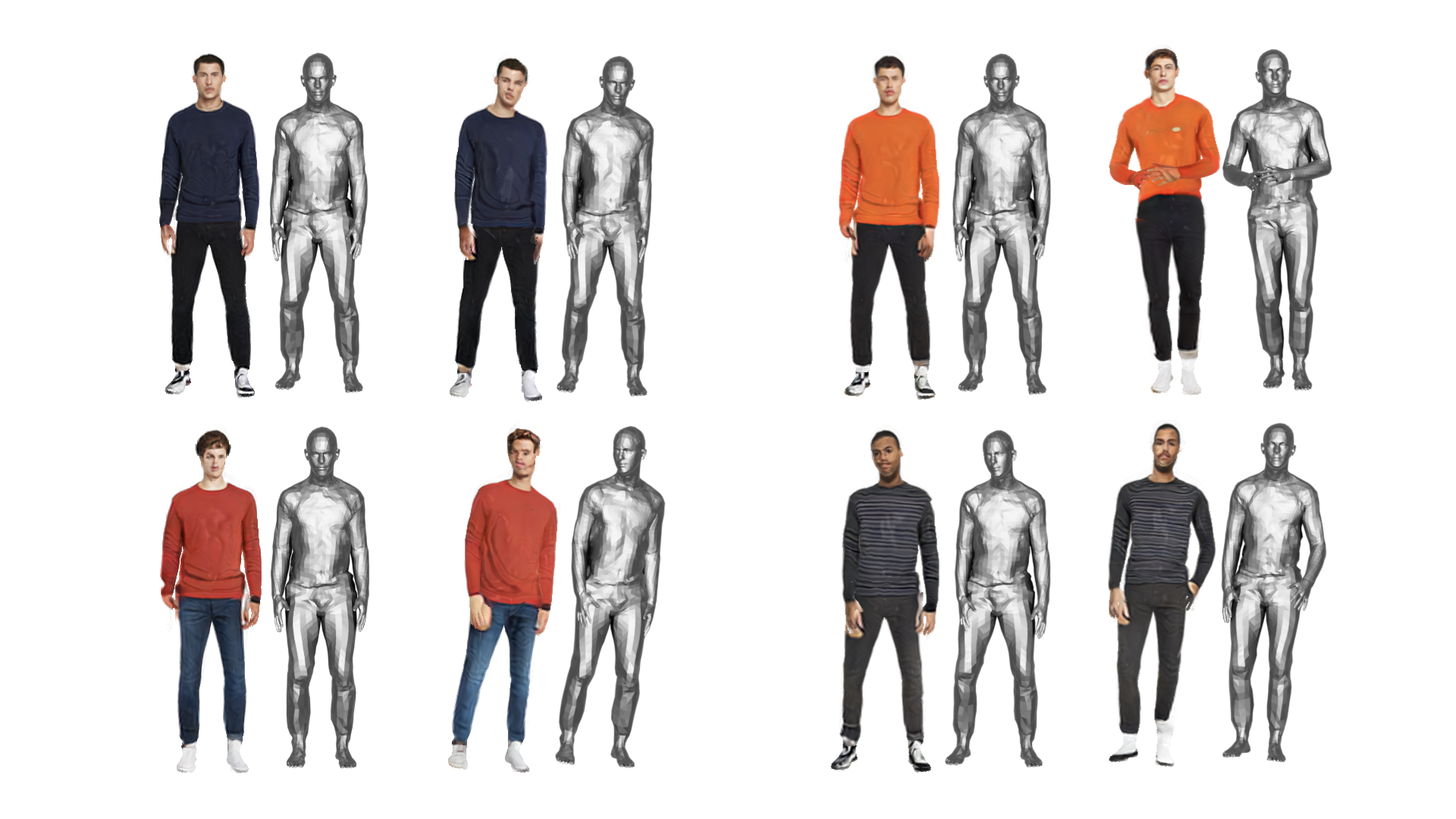}
	}
	\caption{\change{\textbf{Pose control:} Varying pose while keeping other factors fixed. Each row contains two different identities, each in two poses. %
    Texture and geometry meshes are shown side-by-side.}
	}
    \label{fig:pose_control}
\end{figure}

Finally, we combine the generated $\texturemap$ and $\mesh$, and render the textured mesh with a differentiable texture renderer~\cite{ravi2020pytorch3d}. 
We use a global and a patch-based discriminator simultaneously to train $\generatortex$.
While the global discriminator acts on the whole rendered clothed human image,  the patch-based discriminator is a local discriminator \camready{that acts} on the random $64\times 64$ patches extracted from the image.
Following the conditional GAN training scheme, $\clothconditiongeo$ and $\clothconditiontex$ are provided as inputs to both the discriminators.
Empirically, we \camready{find} that the local discriminator helps improve the image quality of the body parts, whereas the global one ensures consistency of the entire structure.

\begin{figure}[t]
	\centerline{
		\includegraphics[width=\linewidth, trim={7cm, 2.0cm, 4cm, 2.0cm}, clip]{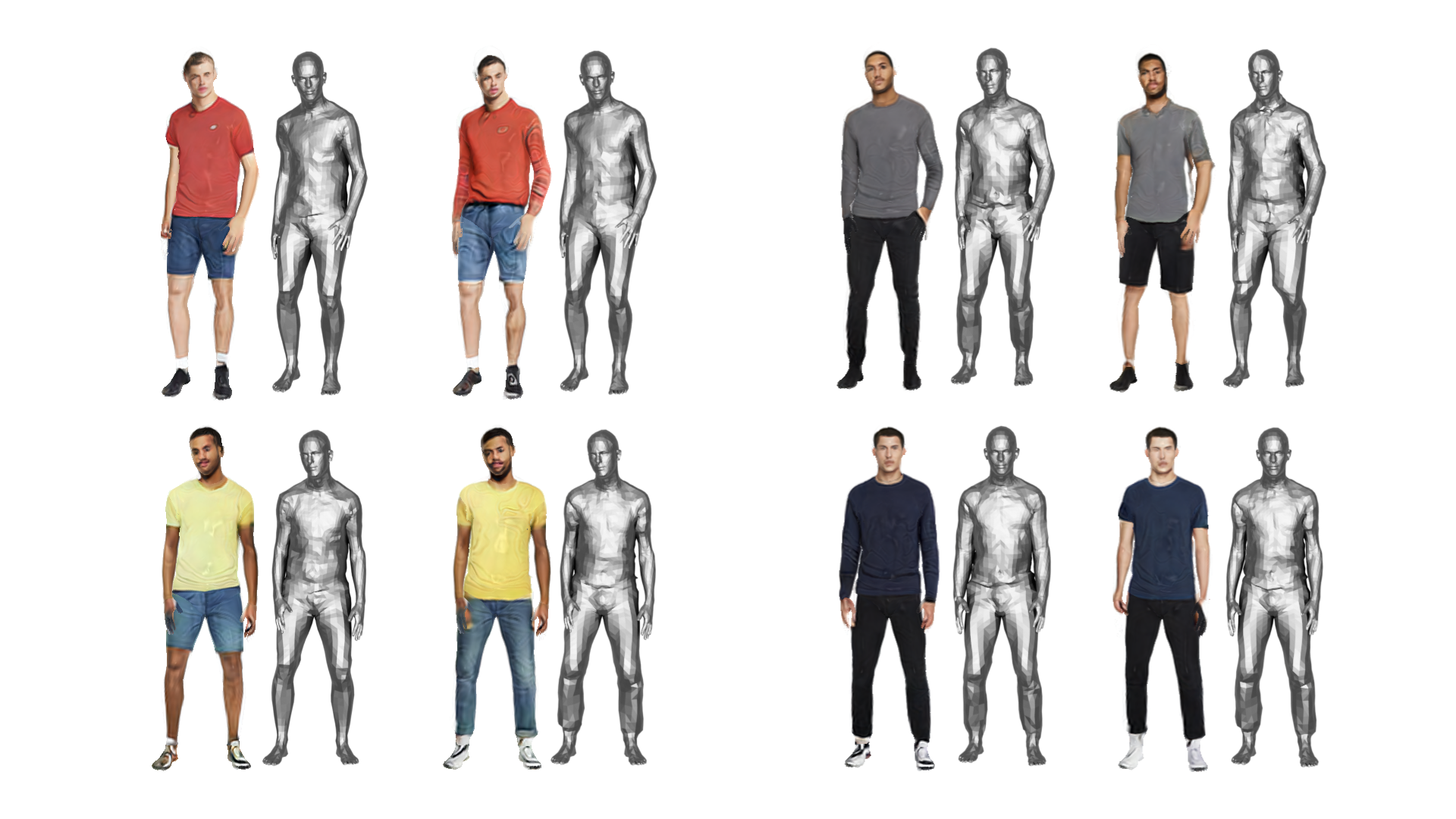}
	}
	\caption{\change{\textbf{Cloth-type control:} Varying $\clothconditiongeo$ while keeping other factors fixed. Each row contains two separate identities. For each identity, we show two different clothing types consecutively. 
    Texture and geometry meshes are shown side-by-side.
    }
	}
    \label{fig:clothing_type_control}
\end{figure}

\subsection{Obtaining clothing texture descriptions}
\label{sec:dataprocessing}

We utilize the language-to-image models CLIP~\cite{radford2021learning} and BLIP~\cite{li2022blip} to automatically label the fashion images with clothing type $\clothconditiongeo$ and clothing color descriptors $\clothconditiontex$.
Specifically, to compute, $\clothconditiongeo$ we pass one fashion image through CLIP~\cite{radford2021learning} and compute its image features.
Then we use augmented prompts for the categorical clothing types, e.g., ``the person is wearing a t-shirt'', ``the person is wearing a shirt'', ``the person is wearing long pants'' etc. as input to CLIP's text feature extractor. 
Finally, we use the scores provided by CLIP for each text input and the corresponding fashion image and select the features corresponding to the text prompt with the highest score as its clothing type descriptor $\clothconditiongeo$.
To compute the color descriptor $\clothconditiontex$, we pass the fashion images to BLIP~\cite{li2022blip} and query its visual question answering (VQA) model with questions such as, ``What is the color of the upper body clothing of the person wearing in the image?''. 
BLIP then outputs a textual description of the clothing color. 
We augment this text with the following sentence, ``The color of the upper body clothing is [BLIP output text] and the color of the pants is [BLIP output text]''.
This is then passed as text input to CLIP to get the text-based feature, which then serves as $\clothconditiontex$.
At inference time, we replace the BLIP generated output with a fixed textual description of the clothing colors, and use this as input to CLIP, to compute $\clothconditiontex$.

\subsection{Training and dataset details}

\methodtitleshort is implemented in PyTorch~\cite{Paszke2019PyTorchAI} and optimized with ADAM~\cite{AdamOptimizer} with a learning rate of 0.001. 
The geometry, texture generators and the discriminators follow the StyleGAN3-t~\cite{Karras2021} architecture. 
The fake examples for the texture discriminators of $\generatortex$ are obtained by rendering the generated clothed body mesh $\mesh$ with the generated texture map $\texturemap$ using PyTorch3D~\cite{ravi2020pytorch3d}.
All generators and discriminators are trained using a non-saturating GAN loss using R1 regularisation~\cite{karras2020analyzing, mescheder2018training} and the adaptive augmentation technique \camready{from} StyleGAN-Ada~\cite{karras2020training}.
The training process follows two stages: 
(1) The geometry generator is trained on the CAPE dataset~\cite{ma2020cape} until the FID converges.
The CAPE dataset~\cite{ma2020cape} consists of SMPL registrations to the scans of 41 subjects wearing different types of clothing.
The dataset comes with six different clothing type variations, namely ``short-short", ``short-long", ``long-long", ``long-short", ``shirt-long" and ``short-short"
Here, the term ``long-short'' represents that the person is wearing a round neck shirt/t-shirt with long sleeves and short pants.
Similarly, the term ``short-long" represents that the person is wearing a round-neck shirt/t-shirt with short sleeves and long pants.
The same terminology is followed for the other labels.
We encode these six clothing types as a 6-dimensional one-hot vector.
Each registered mesh is unposed, i.e., effects of pose articulation and translation are removed, and the vertex offsets from the minimally-clothed SMPL body are represented in the UV space as a displacement map. 
In total, we compute 63069 displacement maps from these registered meshes. 
(2) Then the geometry model is kept fixed and used for training the texture generator until the FID converges. 
\camready{The texture generator} is trained on a curated dataset of fashion images obtained from the catalog images uploaded in the website of Zalando~\cite{zalando}. 
We collected $16362$ fashion images, normalized the images (human centered in the middle) and removed the background.
We run MODNet~\cite{MODNet} on the aligned images to get the segmentation masks for the foreground body and use the pose regressors provided by  ICON~\cite{xiu2022icon} to estimate the SMPL pose and shape of the bodies.
We compute the clothing type and clothing color information utilizing CLIP and BLIP as described in \cref{sec:dataprocessing}.
\camready{More details on the dataset statistics can be found in the Sup. Mat. PDF file.}

\section{Experiments}

\camready{We conduct a three-part evaluation of \methodtitleshort. Initially, we assess its controllability properties. Subsequently, we compare it with state-of-the-art methods. Finally, we execute a detailed series of ablation studies.}

\begin{figure}[t]
	\centerline{
		\includegraphics[width=\linewidth, trim={7cm, 2.0cm, 4cm, 2.0cm}, clip]{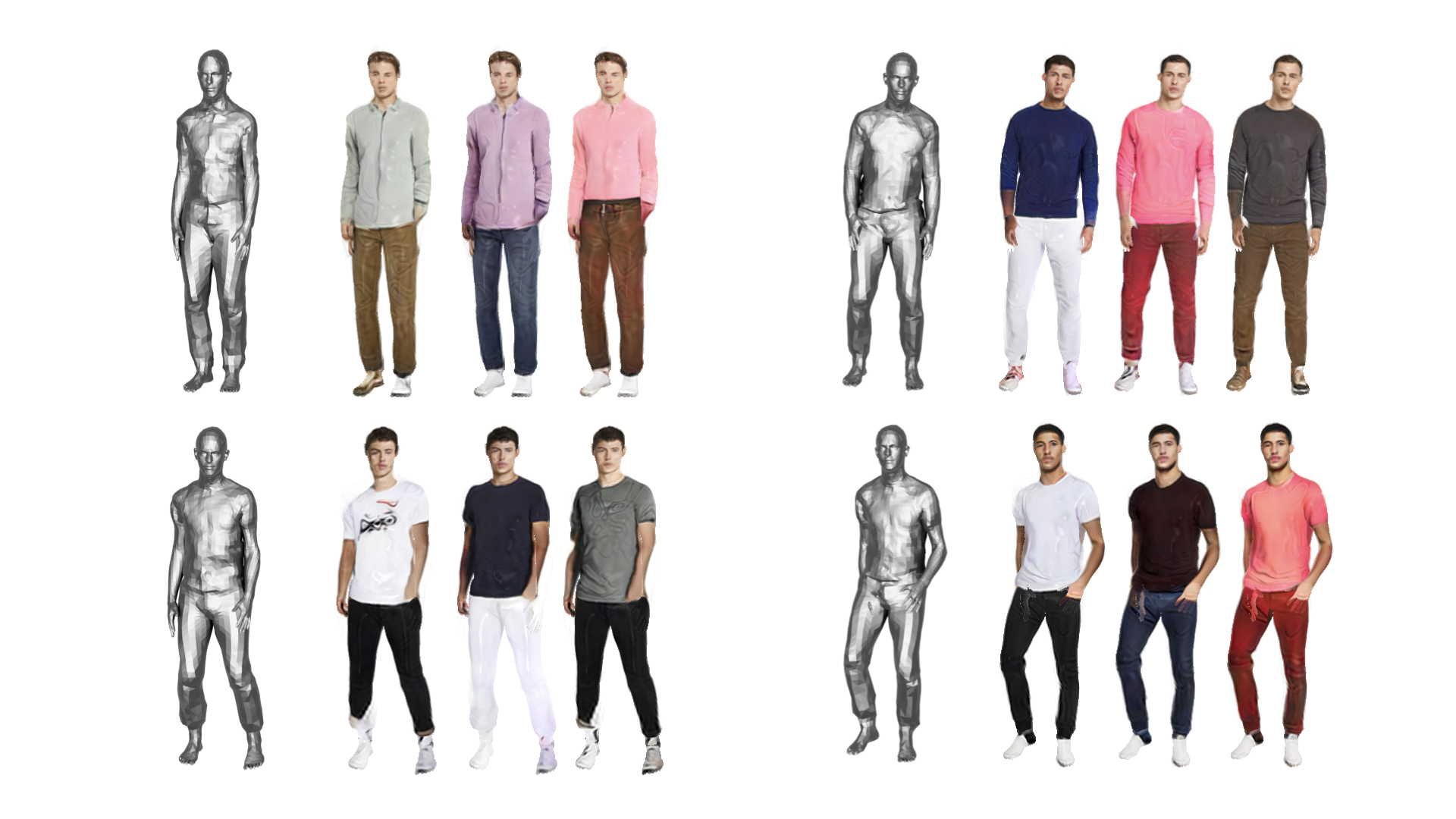}
	}
	\caption{\change{\textbf{Cloth-color control:} $\clothconditiontex$ is varied while keeping other factors fixed. Each row consists of two different clothing geometries and differently colored garments for that geometry.} 
	}
    \label{fig:clothing_color_control}
\end{figure}

\paragraph{Controllability of \methodtitleshort:}
In~\cref{fig:pose_control}, we show generation results by varying the body pose $\posecoeff$, while keeping all other control parameters of \methodtitleshort fixed.
This leads to pose-dependent deformation of the clothing geometry, which is visible in the side-by-side comparison of textured and textureless mesh for each pose-pair of one identity. 
Additionally, we show the pose-dependent deformations only in the geometry in~\cref{fig:pose_dependent_deformations}.
Notice that the appearance of the clothes changes in tandem with the geometry.
In~\cref{fig:clothing_type_control}, we vary the clothing type, $\clothconditiongeo$ keeping all other parameters fixed.
As intended, the clothing type changes from long sleeves and long pants to short sleeves and long pants when $\clothconditiongeo$ is changed accordingly (Row 1, identity 2). 
Fig.~\ref{fig:clothing_color_control} shows that for each clothing geometry, we can generate different colored garments by varying $\clothconditiontex$, fixing the other factors.

\begin{figure}[t]
	\centerline{
		\includegraphics[width=\linewidth, trim={7cm, 19.5cm, 4cm, 2.0cm}, clip]{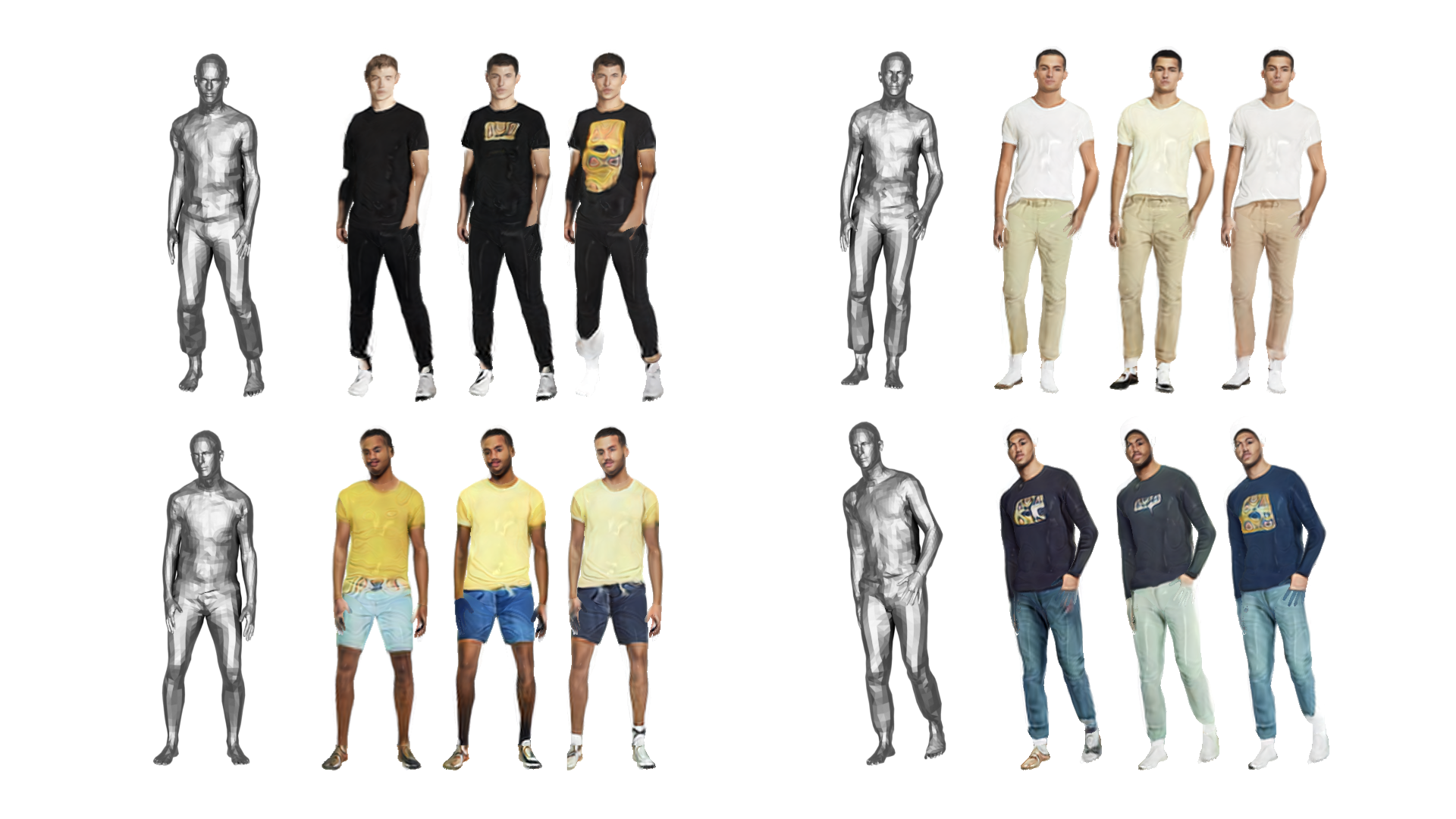}
	}
	\caption{\change{\textbf{Cloth-texture fine control:} Varying $\randomnoise_{tex}$ while keeping other factors fixed. Each row consists of two different clothing geometries, each with textures generated for the same color condition but with different $\randomnoise_{tex}$.}
	}
    \label{fig:clothing_texture_control}
\end{figure}

\begin{figure}[t]
	\centerline{
		\includegraphics[width=\linewidth, trim={7cm, 19.5cm, 4cm, 2.0cm}, clip]{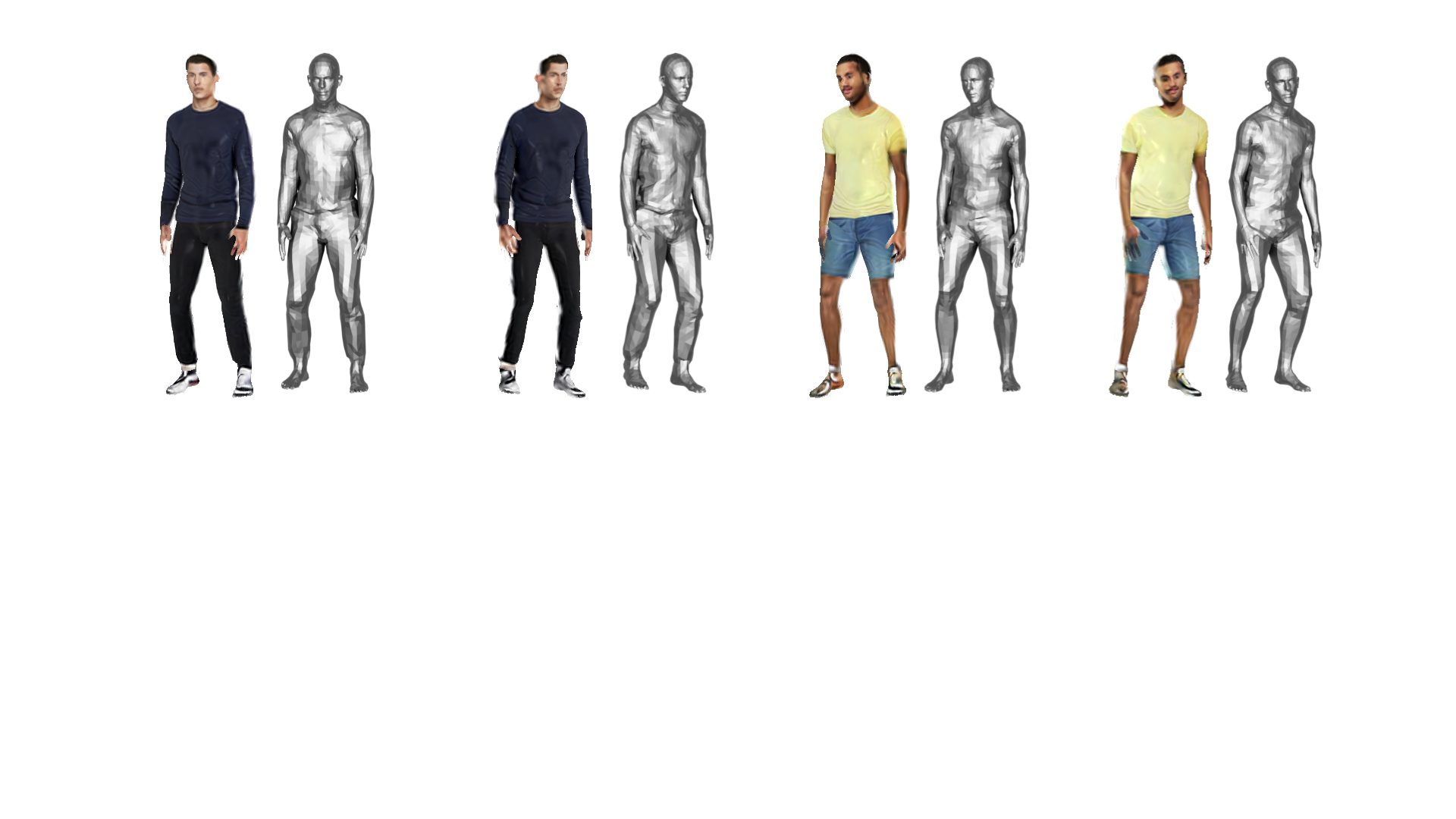}
	}
	\caption{\change{\textbf{Viewpoint changes:} Rotating the textured mesh.
 \TODO{can we show two subjects, and for those only the left and right-most view? Done}
 } 
	}
    \label{fig:clothing_view_control}
\end{figure}

Our model offers further fine-grained \camready{control} over the appearance and allows for fine changes in the texture of the clothing, i.e., it can generate different shades of the same coarse color, and different patterns on the same t-shirt/shirt, etc.
This is achieved by varying $\randomnoise_{tex}$ in~\cref{fig:clothing_texture_control}.
It is worth noting that varying $\randomnoise_{tex}$ changes the texture patterns for identity 1 in row 1 of~\cref{fig:clothing_texture_control} and it generates different shades of the same color.
In contrast to 2D generative modeling, our output is a textured mesh that can be rendered under arbitrary viewpoints, see \cref{fig:clothing_view_control}.

\begin{figure}[t]
	\centerline{
		\includegraphics[width=\linewidth, trim={20cm, 9.5cm, 20cm, 10.0cm}, clip]{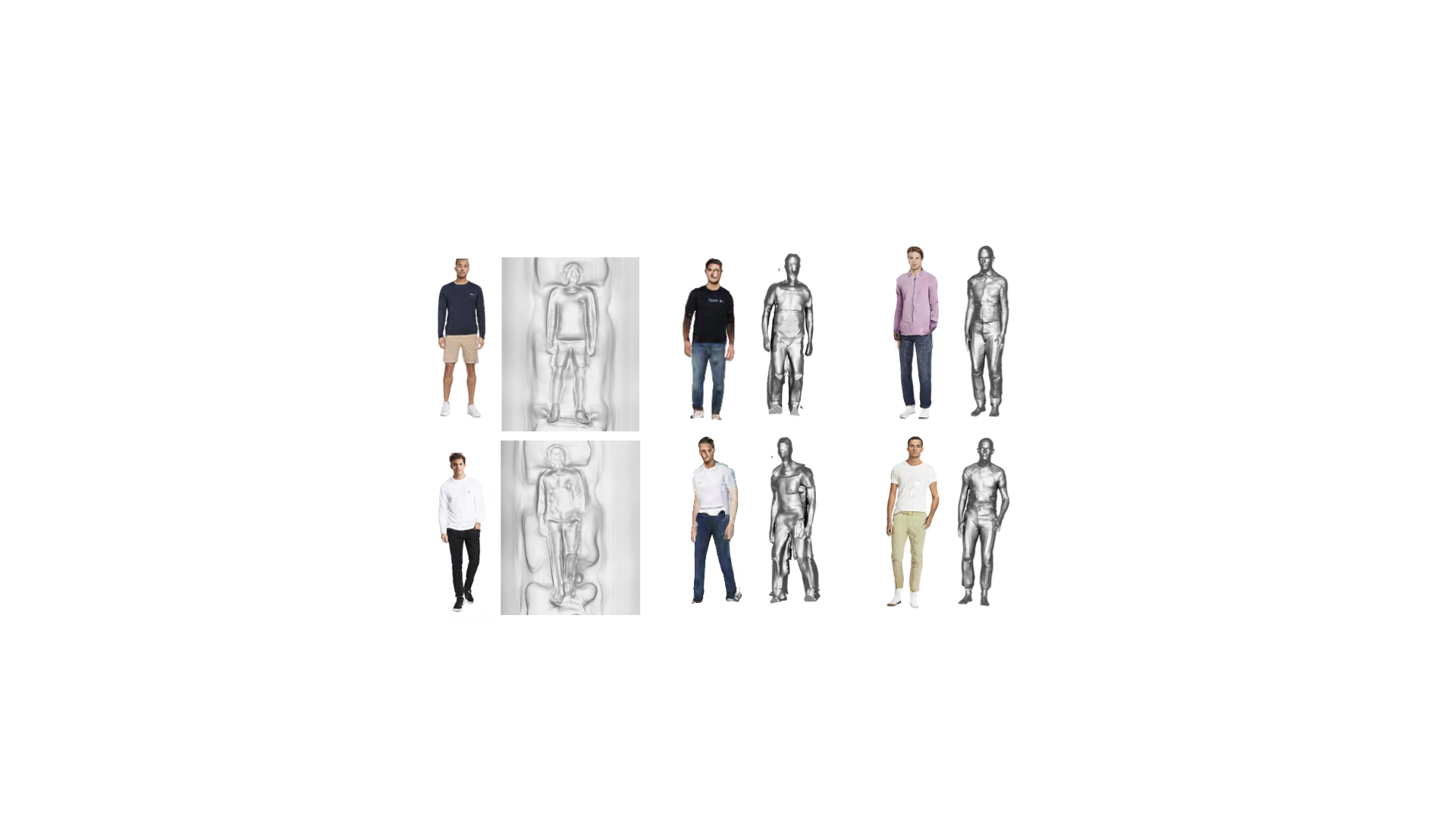}
	}
	\caption{\change{\textbf{Qualitative comparison:} We compare with EG3D (left) and EVA3D (middle). Our rendered humans (right) have comparable quality with EG3D whereas our geometry surpasses both.}
	}
    \label{fig:qualitative_comparison}
\end{figure}

\begin{figure}[]
	\centerline{
		\includegraphics[width=\linewidth, trim={12cm, 12.5cm, 12cm, 11cm},clip]{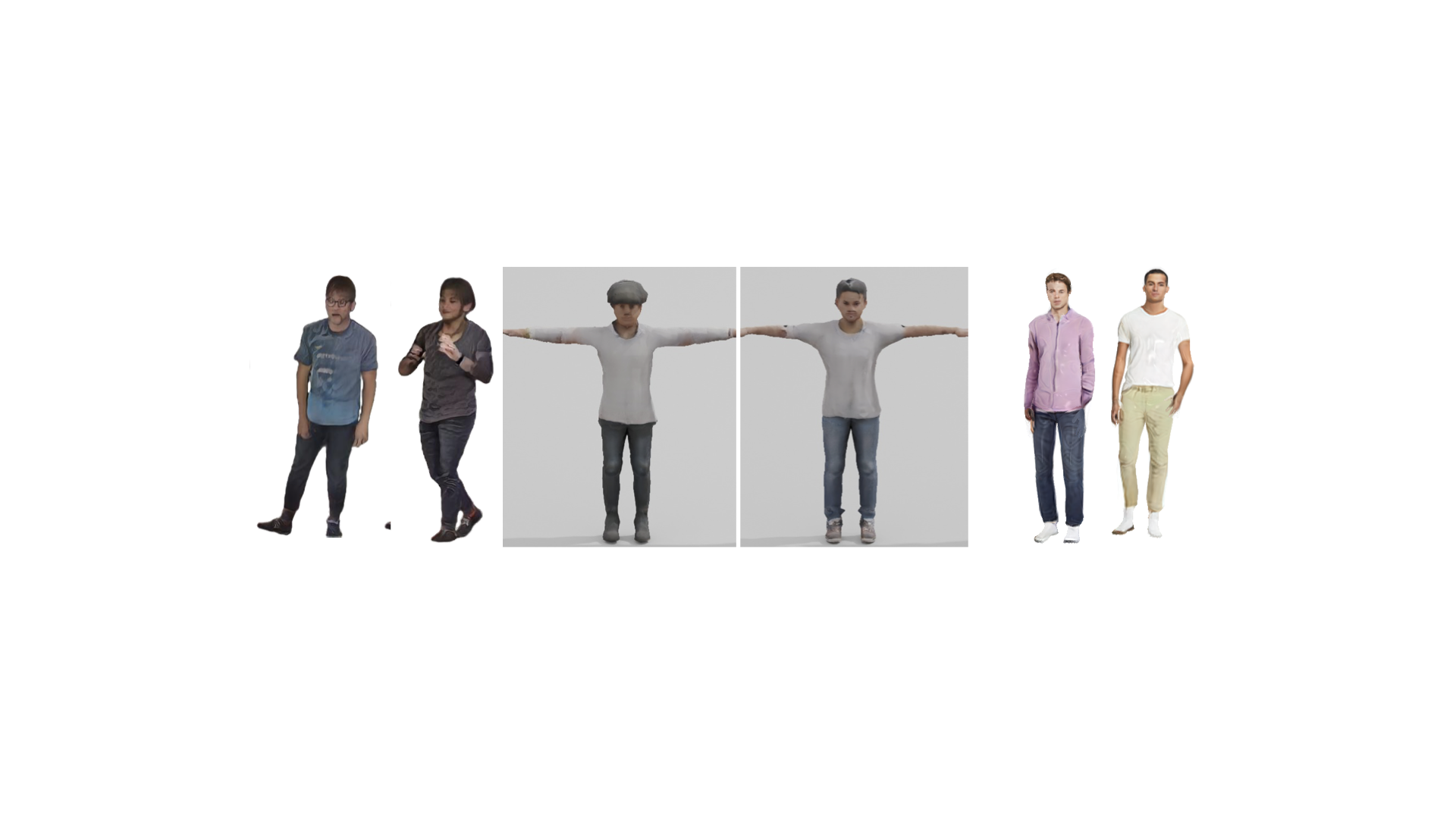}
	}
	\caption{\change{\textbf{Additional qualitative comparisons.} 
 From left to right: StylePeople~\cite{grigorev2021stylepeople}, GET3D~\cite{gao2022get3d}, SCULPT (each with two results). Images are taken directly from the respective publications.}
	}
    \label{fig:additional_qualitative_comparison}
\end{figure}

\paragraph{Comparisons with SOTA:}
We compare \methodtitleshort with the state-of-the-art (SOTA) methods, EG3D~\cite{Chan2022} and EVA3D~\cite{hong2023eva3d} quantitatively in \cref{table:quantitative_comaprison} and qualitatively in \cref{fig:qualitative_comparison}. 
We also provide qualitative comparisons of \methodtitleshort with two additional SOTA methods, namely GET3D~\cite{gao2022get3d} and StylePeople~\cite{grigorev2021stylepeople} in Fig.~\ref{fig:additional_qualitative_comparison}. We compare to these methods only qualitatively because an official implementation of training code of StylePeople~\cite{grigorev2021stylepeople} is not publicly available and GET3D~\cite{gao2022get3d} requires images of people in a canonical pose to train, which is not available for our dataset.

We find that although SCULPT and EG3D both methods generate high-quality images.
However, the underlying geometry\camready{, generated by EG3D,} is of low quality.
We hypothesize that the highly articulated human body is significantly more complex to model as compared to human and animal faces, which have less articulation. 
This renders the training of EG3D \camready{difficult}, leading to undesirable solutions.
Moreover, EG3D does not provide any of the highly desirable control parameters that our model provides. 

Recently, EVA3D~\cite{hong2023eva3d} has been proposed to add controllable articulation to 3D aware generative models of the human body. 
For a fair comparison, we used EVA3D's publicly available implementation and DeepFashion experiment parameters and trained it with our 256X256 texture data. 
\change{Note that the original EVA3D model was trained
on a different dataset with 512X512 resolution, leading to
different results.}
By doing so, EVA3D improves the quality of the generated geometry as compared to EG3D.
But their generated texture is of lower quality \camready{than} SCULPT, which is evident from~\cref{table:quantitative_comaprison} and~\cref{fig:qualitative_comparison}.
Furthermore, as the generated geometry is represented as an implicit function, using it in existing graphics engines requires converting it into meshes, which is time-consuming and often does not preserve the rendering quality. 

\change{
While GET3D's geometric representation can model loose-fitting clothes such as skirts, it lacks articulation and pose control (it generates the body in a canonical pose).
While SCULPT is more limited in topology due to the explicit representation, it enables control over complex, articulated human figures, and generates better textures (\cref{fig:additional_qualitative_comparison}).
Adopting different explicit topologies for different clothing types could greatly enhance the types of \camready{clothes} SCULPT can model. This is left for future work.}

\change{
In contrast to StylePeople,  which employs 2D neural rendering with generated neural textures, our approach estimates accurate clothing geometry that is compatible with standard 3D renderers.
Contrary to StylePeople, SCULPT incorporates a geometry branch. 
The feature outputs from each Style-Block in the geometry generator are added with those from the corresponding ones in the texture generator, resulting in a texture that is consistent with the generated geometry.
This fundamental difference in our architecture allows us to generate better-quality renders compared to StylePeople as can be seen in~Fig.~\ref{fig:additional_qualitative_comparison}.
In summary, SCULPT outperforms the SOTA methods in terms of geometric representation (StylePeople), generated geometry quality (EG3D, EVA3D, StylePeople), articulation (GET3D, EG3D), texture quality (all), and final human image production (all) (\cref{fig:qualitative_comparison} and~\cref{fig:additional_qualitative_comparison}).
}

\begin{table}[]
\centering
\begin{tabular}{lccc}
\hline
Method     & FID $\downarrow$ & KID $\downarrow$ & Precision/Recall $\uparrow$\\ \hline
EG3D   & 7.38  & 0.0036 & 0.79/0.14 \\
EVA3D  & \change{44.11}  & \change{0.0387} & \change{0.30/0.03} \\
SCULPT & \change{9.85} & \change{0.0063} & \change{0.53/0.22} \\ \hline
\end{tabular}
\caption{\textbf{Quantitative comparison:} We evaluate our model using the standard FID, KID, and precision and recall~(\cite{sajjadi2018assessing}) \camready{metrics} against the most recently proposed similar methods. \camready{Our rendering quality is comparable to the state-of-the-art methods.} %
}
\label{table:quantitative_comaprison}
\end{table}

\vspace{-0.5cm}

\begin{figure}[t]
	\centerline{
		\includegraphics[width=\linewidth, trim={8.5cm, 11cm, 10cm, 8.0cm}, clip]{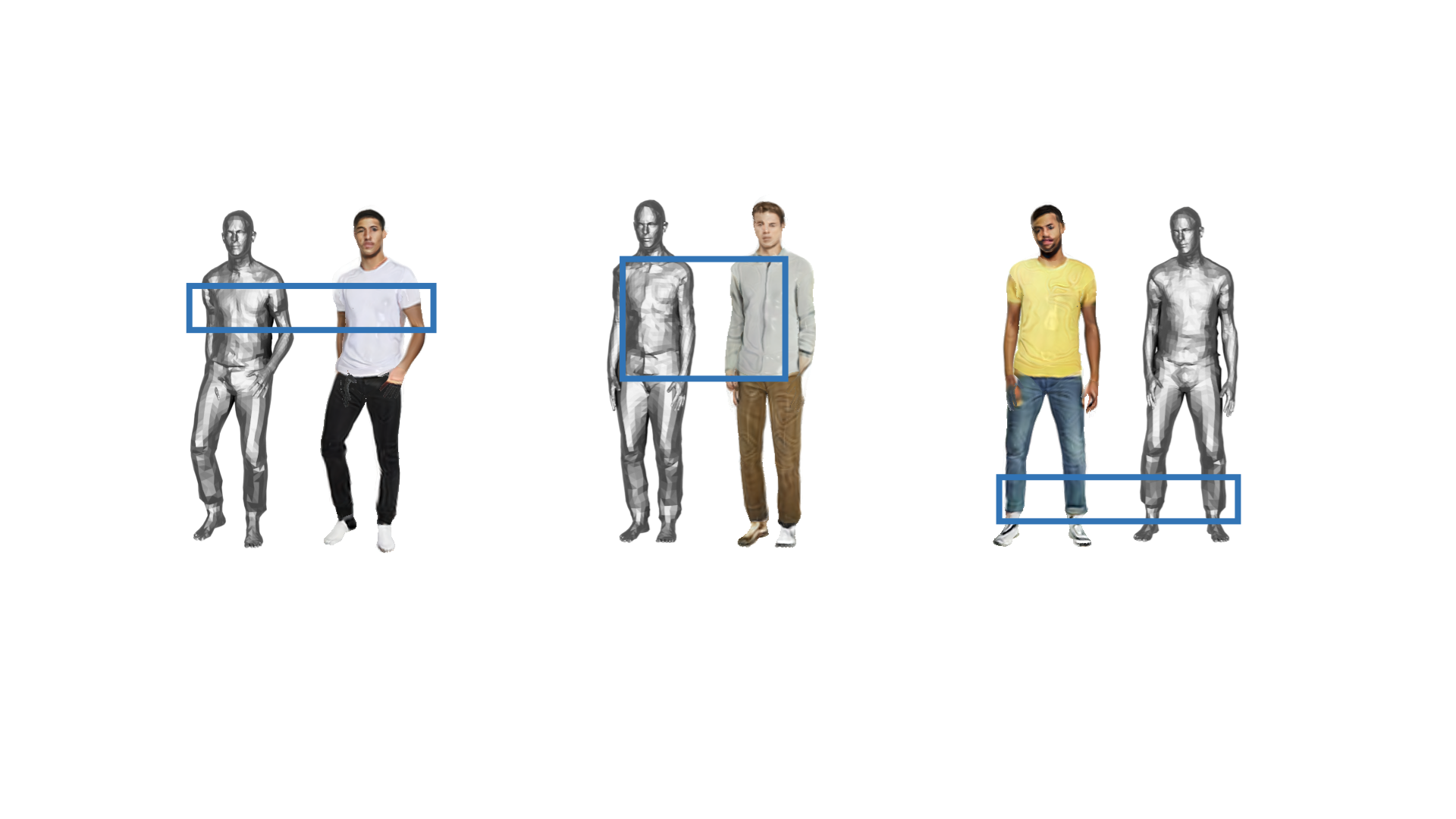}
	}
	\caption{\camready{\textbf{Geometry conforming with texture:} Geometric features like clothing boundaries or wrinkles in different body areas (highlighted in blue boxes) are consistent with texture.}%
	}
    \label{fig:geometry_conform}
\end{figure}

\paragraph{Ablation experiments:}
To better understand the contribution of different components in \methodtitleshort, we perform ablation experiments as shown in~\cref{fig:ablation_study}. 
These experiments involve altering the choice of discriminator combinations, patch sizes, and the conditioning of the texture network by the intermediate activations of the geometry network. 
We train the geometry network conditioned texture network with only one discriminator at a time in cases (b), (c), and (d). 
However, we observe that the global discriminator alone is not capable of generating sharp results (b). 
On the other hand, the patch discriminators work relatively well in improving local parts of the body but lack global correspondence.
Among the local discriminators, we compare the performance of two patch sizes, $64 \times 64$ (d) and $32 \times 32$ (c), and find that $64 \times 64$ performs better. 
\camready{We hypothesize that as the granularity of the local discriminator increases, its field of view diminishes, leading to a higher likelihood of encountering white backgrounds devoid of human elements, particularly in comparison to a $64 \times 64$ patch. This phenomenon potentially impairs the model's overall performance.
Ganokratanaa et al.~\cite{Ganokratanaa2020} observed a similar pattern, where their discriminator using $64 \times 64$ patches outperformed the one with $32 \times 32$ patches.}
We then add the global discriminator along with the local discriminator in cases (e) and (f), which improves the overall performance. 
The $64 \times 64$ patch size also performs best in the combined discriminator strategy as can be seen by comparing (c) and (e), and (d) and (f). 
Finally, we build a baseline where we train the texture generator without conditioning from the geometry network with the dual discriminator strategy, as shown in (a). 
However, the baseline performs poorly compared to our full model (f), as the geometry and texture do not conform to each other. 
\Cref{fig:geometry_conform} shows that the geometry network's conditioning of the texture network allows the texture to conform to the clothing geometry, as observed in the clothing boundaries and wrinkles of the different identities.

\begin{table}[]
\centering
\begin{tabular}{ccccccc}
\hline
\textbf{Ablated} & (a)  & (b)  & (c)  & (d)  & (e)  & (f)  \\
\hline
\textbf{FID $\downarrow$}            & 28.2 & 31.6 & 24.5 & 16.1 & 19.2 & 9.85 \\ \hline
\end{tabular}
\caption{\textbf{Ablation study:} (a) full model without geometry conditioning; (b) full model trained with only global discriminator; (c) full model trained with only local discriminator of patch size $32\times32$; (d) full model trained with only local discriminator of patch size $64\times64$; (e) full model trained with both global and local discriminator of patch size $32\times32$; (f) full model trained with both global and local discriminator of patch size $64\times64$.}
\label{fig:ablation_study}
\end{table}

\begin{figure}[t]
        \vspace{-1em}
	\centerline{
		\includegraphics[width=\linewidth]{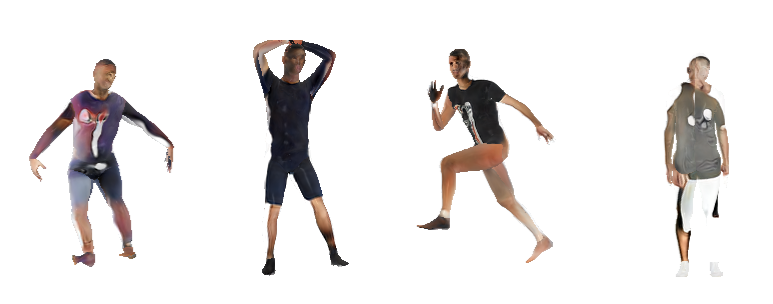}
	}
        \vspace{-1em}
	\caption{\textbf{Limitations:} 
    The texture quality degrades for out-of-distribution poses (columns 1-3) and for the body's back (right).
	}
    \label{fig:ood_pose}
\end{figure}

\paragraph{Limitations and discussion:}
\camready{Similar to existing generative clothed human body models, \methodtitleshort generates poor-quality textures in the backside region because of a dataset bias towards frontal and near-frontal views, and to ``typical'' fashion poses in the fashion image dataset used for training. 
The dataset bias can be observed from the dataset statistics provided in the \textbf{Sup. Mat. PDF file}.
The quality also degrades for challenging, unseen body poses~\cref{fig:ood_pose}. }
The existing datasets further lack diversity in \camready{age}, race, skin tone, and gender. 
The predominance of male examples is due to the CAPE dataset bias towards tight clothing and the prevalence of male subjects in our texture training data.
\camready{To overcome the limitation in view- and pose diversity, one can train our model on multi-view datasets or videos of subjects in the same clothing but in varying poses and visible from different viewpoints.}
\methodtitleshort sometimes generates hand textures with the same color as the clothing. 
\camready{The issue partly arises from about 30\% of the training set's fashion images showing hands in pockets, as observed from manual image examination of a hundred images. 
However, the CAPE dataset used for geometry training lacks hands-in-pockets instances, leading to model ambiguity in recognizing hand positions in pockets. 
Augmenting the dataset with annotations specifying hand positions could mitigate this.}
\change{
Finally, our model also has topological limitations in modeling loosely fitting clothes such as skirts or long dresses.
To handle loosely fitting clothing such as skirts, our method would require a corresponding mesh template that offers the correct topology.
Given the limited range of typical clothing typologies, it could be feasible to design a few different typologies to accommodate such clothing types but this is outside of the scope of this work.
Instead, we focus on developing a novel method for learning geometry and texture without paired 3D training data.
We use SMPL as a template for our explicit representation (mesh) which is compatible with current graphics engines.
}

\section{Conclusion}

We have introduced \methodtitleshort , a generative model that creates 3D virtual humans using explicit geometry (mesh) and appearance (texture maps). \methodtitleshort  represents clothing geometry as offsets from the SMPL body's vertices, and includes a generative model for texture maps based on clothing type and appearance. This approach effectively combines traditional graphics elements such as meshes, forward rendering and texture maps with modern 3D-aware generative models.
Our texture model is trained from unpaired 2D-3D data, making it easy to use and retrain on new data. The clothing geometry is learned using a dataset of 3D meshes.%
The trained model offers generative control over clothing geometry and appearance, with semantic control over clothing type and color, while still retaining SMPL's pose articulation and body-shape variation.
Compared to previous 3D-aware generative models, \methodtitleshort offers greater control and produces higher quality geometry and textures.

\qheading{Acknowledgments:}
The authors would like to thank Nikos and Joachim for their valuable tips on Blender renderings, Benjamin for IT support, and Tsvetelina, Taylor, and Tomasz for their support in conducting the perceptual study. 

\qheading{Disclosure:}
\url{https://files.is.tue.mpg.de/black/CoI_CVPR_2024.txt}

{
    \small
    \bibliographystyle{ieeenat_fullname}
    \bibliography{main}
}
\clearpage
\setcounter{page}{1}
\maketitlesupplementary

\begin{figure}[t]
	\centerline{
            \includegraphics[width=\linewidth]{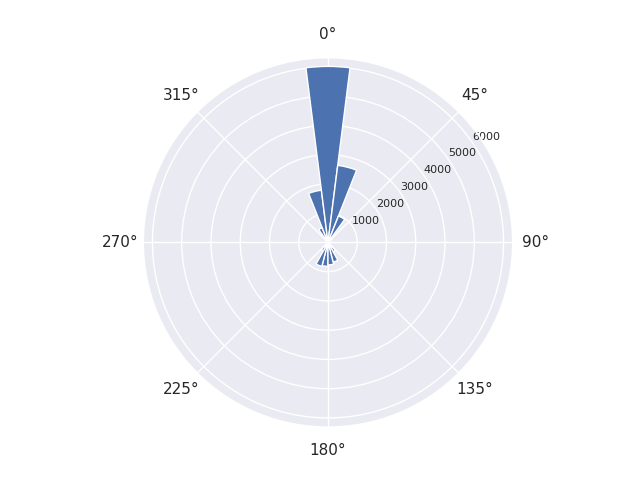}
	}
	\caption{Histogram of the body rotations of the used training corpus with respect to the camera view (0° is frontal). %
	}
    \label{fig:clothing_type_stats_view_stats}
\end{figure}

\section{Dataset statistics:} 

Refer to~\cref{fig:clothing_type_stats_view_stats} for the view statistics of the \textit{SCULPT dataset}. 
It is observed that the dataset exhibits a bias towards both frontal and near-frontal views. 
The dataset offers a variety of clothing types, including `short sleeve T-shirt/short trouser'', ``short sleeve T-shirt/long trouser'', ``long sleeve T-shirt/long trouser'', ``long sleeve T-shirt/short trouser'', ``shirt/long trouser'', and ``shirt/short trouser'' which are similar to the assortment found in the CAPE dataset. 
The labels were automatically generated using CLIP, as detailed in Sec. 3.2. %
The dataset contains 2,483 ``short-short'', 6,260 ``short-long'', 335 ``long-short'', 3,425 ``long-long'', 939 ``shirt-short'', and 2,920 ``shirt-long'' items, where ``short-short'' refers to ``short sleeve T-shirt/short trousers'', and so forth. 
Regarding the color types in the training dataset of fashion images, there are descriptions for 115 different colors. 
Examples of these colors include red, blue, green, khaki, pink, peach, and tan, among others. 
We plan to release the dataset annotations for research purposes.

\section{BLIP texture description accuracy:}

In a perceptual study with 2000 labeled images on Amazon Mechanical Turk, BLIP labels were judged to be correct  92.7\% of the time for upper body clothing and 89.7\% for lower body clothing. During the study, participants received images alongside associated BLIP labels and assessed their validity with a ``yes/no" answer. We treat the human judgements as ground truth.
The common point of mismatch between the participants and BLIP labels was nearby colors like khaki or tan etc.

\section{Parameters of the differentiable rendering:} 
Our model utilizes PyTorch3D's soft rasterizer for differentiable rendering, with a zero blur radius for one-to-one pixel-triangle correspondence. A directional light with fixed intensity and orthographic projection is employed for mesh rendering. Body orientation or view for each rendered image is randomly chosen from the training dataset, with this randomization applied per image in each batch during generator forward passes. Check our training and inference codebase for reproducibility.

\end{document}